# 基于空地障碍分布模式匹配的火星巡视器空间定位与地面验证

周朗，张志泰，王虹量

（同济大学测绘与地理信息学院，上海 200092）

**摘 要**：地外天体巡视器空间定位是保障巡视器大范围探测的前提。美国 Mars 2020 火星探测任务中首次搭载无人机，其能够获取火星表面局部的高分辨率地形，从而可以通过将巡视器影像与无人机影像进行匹配的方式以对巡视器进行空间定位。但火星表面往往纹理贫乏，且无人机影像与巡视器地面影像间存在着较大的成像视角差异，两者直接进行匹配难度较大。据此，本文提出了一种基于空地障碍分布模式匹配的火星巡视器空间定位方法，通过建立巡视器影像与无人机基准地图中的石块分布模式间的对应关系，获得巡视器在无人机基准地图中的空间位置。我们使用本文方法在同济大学月球与深空探测精密测绘综合实验场开展了巡视器定位地面试验，实验结果验证了该方法的可行性，为后续我国开展类似工程任务提供参考。

**关键词**：巡视器定位；空地影像匹配；障碍识别；深空探测；火星无人机

**Abstract:** Rover localization is one of the perquisites for large scale rover exploration. In NASA's Mars 2020 mission, the *Ingenuity* helicopter is carried together with the rover, which is capable of obtaining high-resolution imagery of Mars terrain, and it is possible to perform localization based on aerial-to-ground (A2G) imagery correspondence. However, considering the low-texture nature of the Mars terrain, and large perspective changes between UAV and rover imagery, traditional image matching methods will struggle to obtain valid image correspondence. In this paper we propose a novel pipeline for Mars rover localization. An algorithm combing image-based rock detection and rock distribution pattern matching is used to acquire A2G imagery correspondence, thus establishing the rover position in a UAV-generated ground map. Feasibility of this method is evaluated on sample data from a Mars analogue environment. The proposed method can serve as a reliable assist in future Mars missions.

## 1 研究背景

地外天体巡视器就位探测是深空探测的重要手段，巡视器空间定位是大范围巡视探测的前提条件。火星等地外天体不存在地球上已得到广泛应用的 GNSS 系统，巡视器无法通过卫星获得位置信息。常用的地外天体巡视器定位方法主要包括航迹推算[1]、视觉里程计[2]、惯性导航[3]等。但此类方法的定位误差通常会随巡视器行驶里程增加而累积，从而导致长距离下的定位漂移。

美国国家航空航天局在其 Mars 2020 火星任务中首次随毅力号火星巡视器搭载了一架名为机智号的无人机[4]，其设计最长飞行距离约 300 米，飞行高度约 3-10 米，可获取火星表面局部的高分辨率影像，从而为巡视器定位提供了新的思路，即以无人机影像为中介，通过将无人机影像与卫星影像和巡视器影像分别进行配准，以确定巡视器

在卫星全局地图中的绝对位置。其中的关键问题即在于，如何对存在较大成像视角差异的无人机影像与巡视器影像进行空地匹配。

据此，本文提出了一种基于空地障碍分布模式匹配的火星巡视器空间定位方法，主要内容包括：1）利用无人机影像生成局部基准地图，包括三维点云及 DOM 影像；2）分别从无人机 DOM 影像与巡视器影像中对石块障碍进行提取；3）通过巡视器立体影像匹配获取石块障碍在巡视器相机坐标系下的坐标，从无人机三维点云中获取无人机 DOM 影像石块障碍在全局坐标系下的位置；4）建立 DOM 影像中石块分布与巡视器影像中石块分布间的对应关系，进而获取巡视器相机在无人机基准地图中的空间位置。我们使用地面模拟数据对这一方法进行了巡视器定位地面测试，测试结果验证了本文方法的可行性。

## 2 实验数据

本文所使用的实验数据均采集于同济大学月球与深空探测精密测绘综合实验场，实验场内建有火星地表环境模拟场，具有多尺度障碍等模拟地形场景。为获取实验场局部基准地图，本文使用一台大疆无人机对实验区域进行了高分辨率成像，共获取 68 张无人机影像。此外，利用自行搭建的双目相机系统模拟巡视器视角成像平台，在不同位置对实验区域进行了立体相对数据采集，立体相机影像均经过标定校正。实验拍摄的示例无人机影像和地面影像如图 1 所示。在试验场内均匀采集了控制点以将无人机基准地图转换到 WGS-84 坐标系下，同时利用 RTK 获取了相机成像时刻的 WGS-84 坐标，以对最终的定位结果进行定量评价。

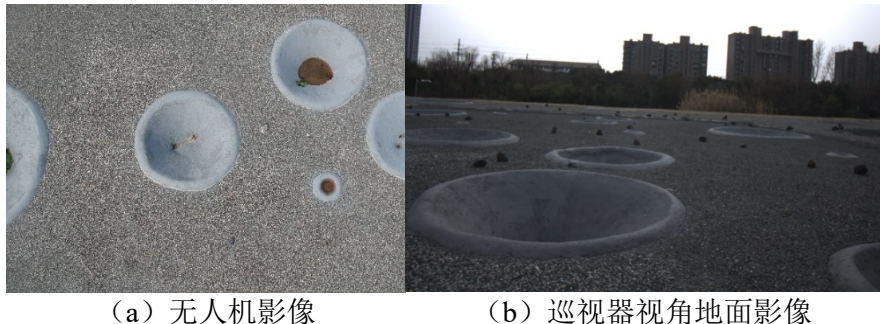

（a）无人机影像　　（b）巡视器视角地面影像

图 1 实验数据示意图

## 3 研究方法与实验结果

从图 1 可以看出，实验场内除了零星分布的石块及撞击坑等障碍，其他区域纹理较为贫乏，且由于无人机影像与地面影像成像视角差异较大，两者直接进行匹配不易成功。图 2 为实验场局部区域空地影像 ASIFT 算法[5]匹配的结果，可以看出在此类低纹理且大成像视角差异的环境下，传统匹配方法难以得到正确匹配结果。

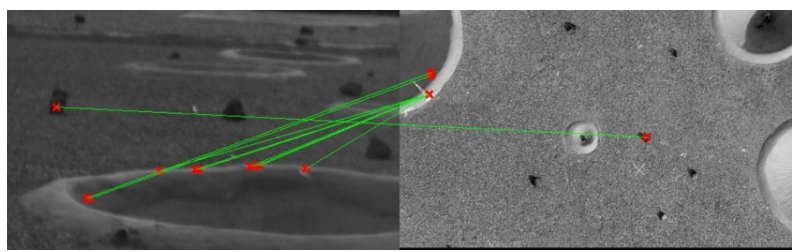

图 2 实验场局部区域 ASIFT 算法匹配结果

据此，本文提出了一种基于空地障碍分布模式匹配的火星巡视器空间定位方法，其主要流程如图3所示。

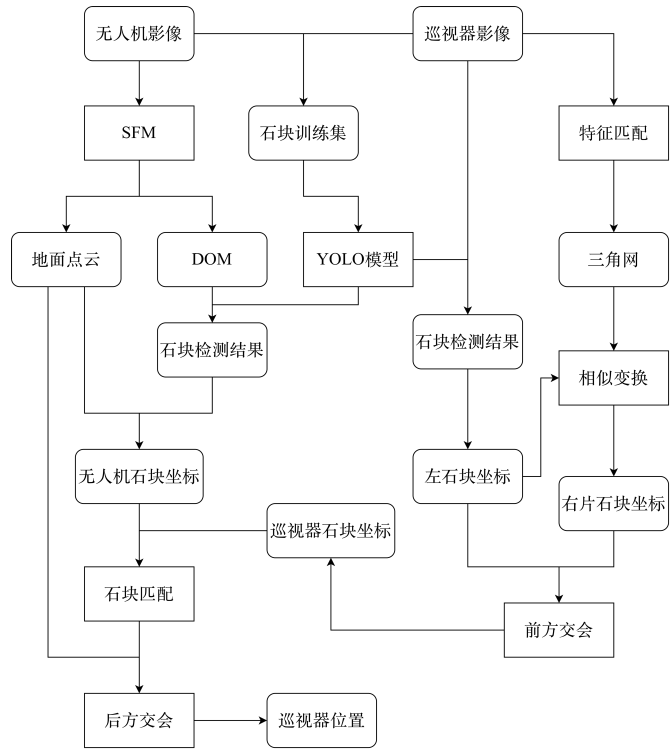

图 3 空地协同定位流程示意图

### 3.1 无人机基准地图构建

运动结构恢复（Structure From Motion, SFM）是一种使用多张不同视角图像对场景进行三维重建的方法，已被广泛应用于计算机视觉领域的多种任务中。相比于传统摄影测量学科中的解析空中三角测量方法，SFM 方法对输入数据质量要求较低，且可得到精度相当的结果[6]，具有更强的稳健性。本文研究使用 SFM 方法对无人机影像序列进行处理，利用 UAV-Mapper 软件[7]生成实验区域的三维点云及 DOM 影像，作为后续空地匹配的基准地图，其结果如图 4 所示。

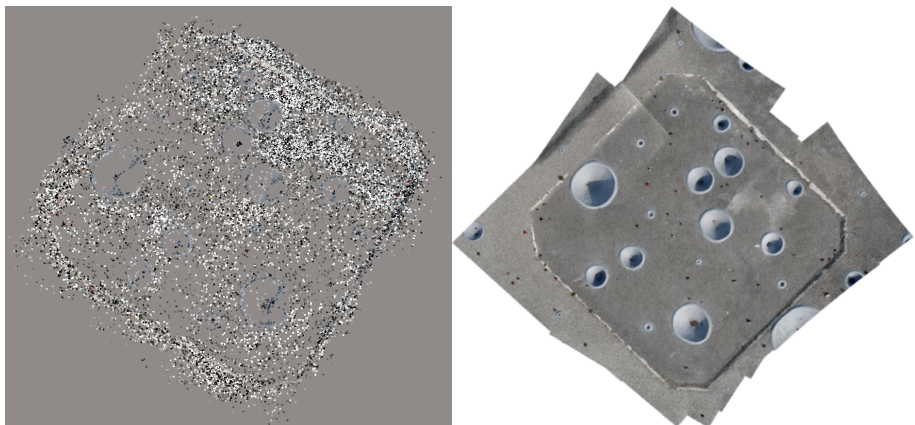

（a）三维点云　　　　　　（b）DOM 影像

图 4 实验场无人机基准地图

## 3.2 空地影像石块提取

本文以无人机 DOM 影像与地面立体影像作为数据源，使用 YOLOv5[8]目标检测方法分别对空地影像中的石块进行提取。该方法将边界框回归和目标分类合并在一步进行，同时在输入端引入了自适应锚框计算等算法优化，在提升检测速度的同时具有较好的检测效果。

在目标检测模型的训练方面，我们使用拍摄的多幅无人机影像与地面影像构建训练数据集。首先对原始图像中的石块进行手动标注，随后将每张图像裁切为 4*4 的子影像，并分别重采样至 640×640 尺寸，按 70%-20%-10%的比例进行训练-验证-测试集的划分，最终获取的用于训练的图像共计 1888 张。

综合考虑算力和检测效果，我们选用了 YOLOv5m6 预训练模型作为训练的初始权值，共训练 300 个 epoch，取训练过程中 mAP 指标最高的一次权值作为训练结果。最终训练得到的模型在测试集上的主要指标如表 1：

表 1 YOLOv5m6 模型石块测试集检测效果

| Precision | Recall | mAP@.5 |
| --- | --- | --- |
| 0.906 | 0.891 | 0.913 |

利用训练得到的模型分别对无人机 DOM 影像和选择的巡视器视角影像进行检测，其结果如图 5 所示。石块检测结果为若干个矩形目标框，为简化后续流程，我们将检测到的石块位置定义为所对应的目标框中心像素。

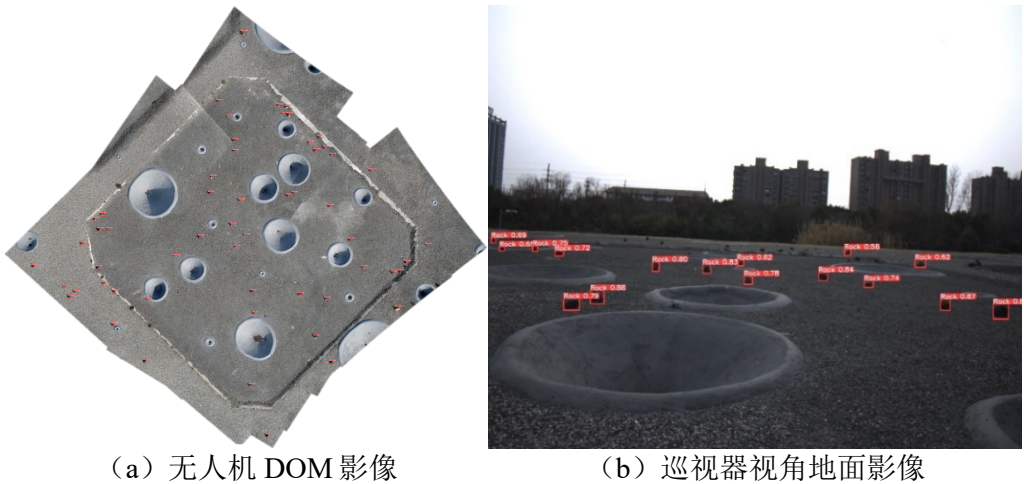

（a）无人机 DOM 影像　　　　（b）巡视器视角地面影像

图 5 石块检测结果

## 3.3 空地影像石块三维点集生成

针对无人机 DOM 影像中的石块，直接在其对应的三维点云中查询获取其三维坐标，进而生成无人机影像石块三维点集。为保证石块三维坐标精度，本文仅仅使用了地面影像中距离相机较近的石块用于三维点集生成。针对地面影像中的石块，可以通过前方交会获取石块点在左相机坐标系下的三维坐标，但其前提是能够有效确定左相机影像中的石块点在右相机影像中的同名点，其本质是一个密集匹配问题。

本文使用了一种基于狄洛尼三角网变换的影像匹配方法，其具体流程如下：（1）使用 ORB-SLAM2 框架[9]中的特征提取与匹配方法，获取立体影像间的稀疏同名特征点，其原理为通过一维核线搜索确定左右影像 ORB 特征点间同名关系，并通过面片相关性对匹配结果进行优化；（2）利用左影像的特征点构建狄洛尼三角网；（3）选定左

影像中任一石块点 $P_i$，确定其所在的三角形 $T_i$；（4）获取三角形 $T_i$ 的三个顶点 $V_{ij}$ 及其在右影像上同名点 $V'_{ij}$；（5）计算点集 $V_{ij}$ 与 $V'_{ij}$ 间的仿射变换关系 $f_i$，将 $f_i$ 施加于石块点 $P_i$ 上，据此得到其在右相机影像上的同名点 $P'_i$；（6）使用前方交会方法计算石块同名点对 $(P_i, P'_i)$ 在以左相机为中心的相机坐标系中下的三维坐标；（7）重复步骤3~6，至所有石块坐标均计算完毕，最终生成地面影像石块三维点集。

将获取的空地影像石块三维点集分别投影在二维平面上，得到如图 6 所示的石块障碍分布图。

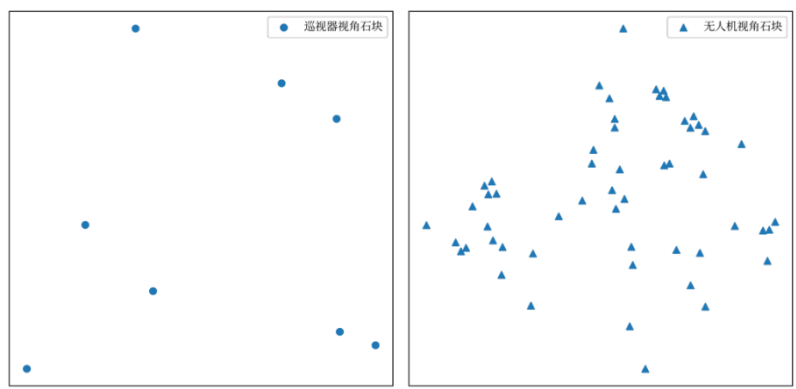

（a）巡视器视角石块分布　　（b）无人机视角石块分布

图 6 空地影像石块障碍分布图

## 2.4 石块分布模式匹配

考虑到地面相机成像视角近似平行于地面，而无人机成像视角则可视作垂直摄影，可以仅考虑两者在二维平面上的关系。本文利用平面仿射变换对空地石块分布模式间的关系进行建模。令无人机平台提取到的石块坐标为 $X, Y$，巡视器视角下的石块坐标为 $x, y$，则两者之间的变换关系可表示为：

$$\begin{bmatrix} X \\ Y \\ 1 \end{bmatrix} = \begin{bmatrix} a_1 & b_1 & c_1 \\ a_2 & b_2 & c_2 \\ 0 & 0 & 1 \end{bmatrix} \begin{bmatrix} x \\ y \\ 1 \end{bmatrix} \tag{1}$$

本文使用文献[10]提出的"试探匹配加一致性验证"的方法进行分布模式匹配，其基本思想为每次随机抽样石块对计算变换参数，计算该变换模型下所有坐标变换结果，按照最近邻的原则确定同名石块对并计算整体残差，最终取多次抽样中残差最低的一次作为匹配结果。由空地平台石块坐标估计得到的变换关系如图7所示。

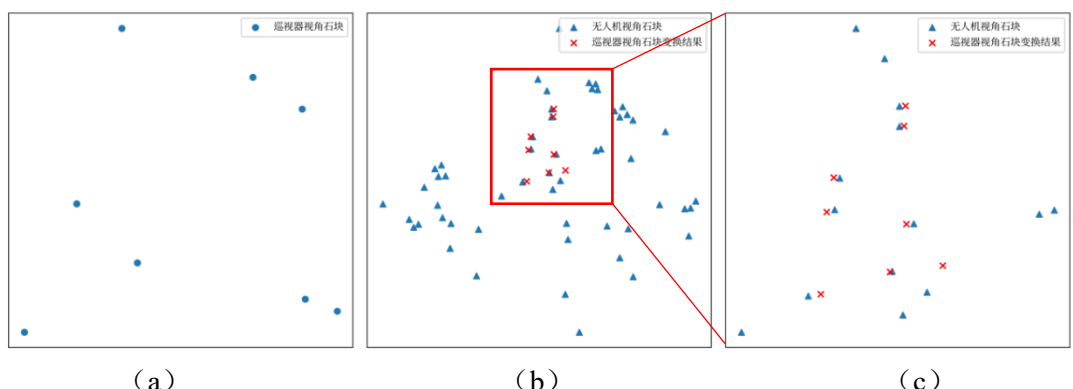

（a）　　　　　　　　（b）　　　　　　　　（c）

图 7 空地石块分布模式匹配结果。（a）巡视器视角石块分布；（b）巡视器石块坐标经仿射变换模型变换结果与无人机石块坐标对比；（c）放大图

依据该变换关系所确立的空地影像同名关系如图8所示。

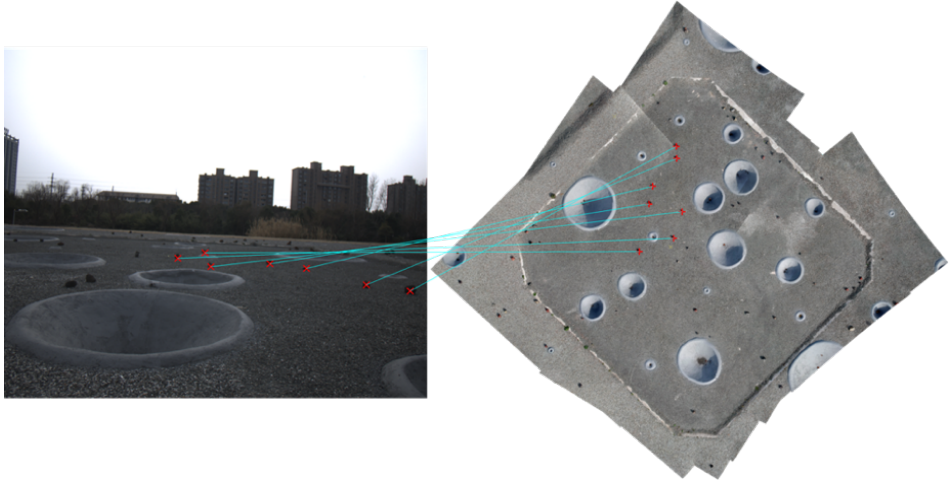

图8 地面影像（左相机）与无人机DOM影像间石块同名关系

### 3.5 空间后方交会定位

经过上述石块分布模式匹配流程，可以得到空地影像间石块位置的同名点对关系，进而可使用空间后方交会算法计算巡视器位置。

相较于航空摄影影像，巡视器拍摄影像一般具有大倾角特性，这使得传统基于初始值进行迭代的空间后方交会方法[11]未必适用于本研究。本文使用一种改进的空间后方交会算法[12]，其在无初始值条件下可保证全局收敛性。该方法使用四元数而非传统方法中的欧拉角来描述相机姿态，并以绝对定向和正交投影变换模型替代了共线方程模型，构造相机位置$r_s$、投影点$r'$、方向矢量$v$，旋转矩阵$M$间的约束关系如下：

$$\begin{cases} r' = Vr' \\ M(r - r_s) = VM(r - r_s) \end{cases} \quad (2)$$

式（2）中，$V = \frac{vv^T}{v^T v}$为投影矩阵。对于所使用的的$n$个控制点，建立损失函数：

$$E(M, r_s) = \sum_{i=1}^{n} \|(M^T M - V_i)M(r_i - r_s)\|^2 \quad (3)$$

按照此方法进行空间后方交会即为求解使$E(M, r_s)$最小的$r_s$，将式（3）展开后使用极值定理求导可解得：

$$r_s(M) = \left(\sum_{i=1}^{n} M^T H_i M\right)^{-1} \sum_{i=1}^{n} M^T H_i M r_i \quad (4)$$

式（4）中，$H_i = (M^T M - V_i)^T (M^T M - V_i)$。当$M$已知时，$r_s$为满足约束条件的最优解。由于本文所实现的为平面定位，故保留组成$r_s$的三个线元素$X, Y, Z$中的$X, Y$即可。

该方法使用正交投影模型构建约束条件，无需对方程进行线性化而可直接迭代求解非线性方程，故对相机姿态有良好的适应性，较适合于火星巡视器的应用场景。

据此，可以计算左相机的空间位置如图9所示。

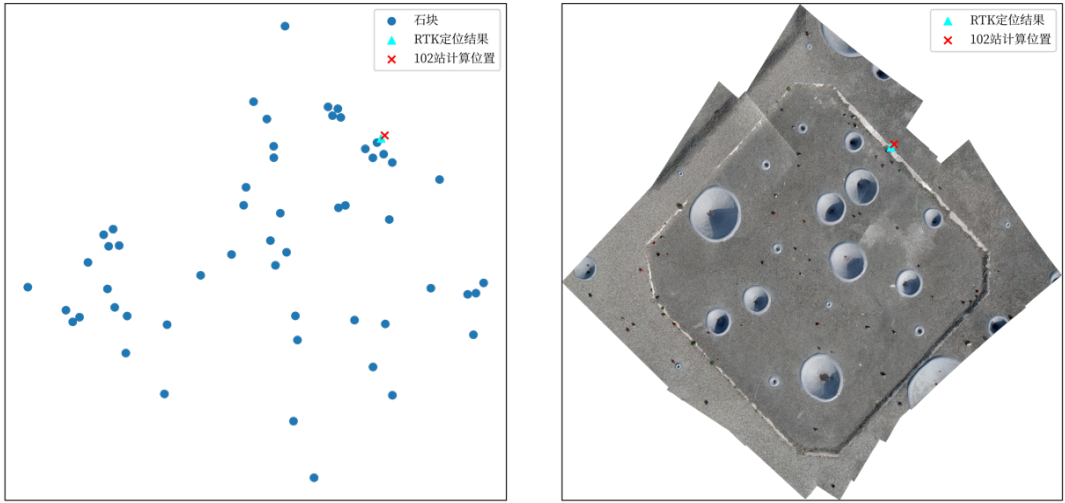

|  (a) 本文方法定位结果 | （b）定位结果叠加 DOM |

图 9 巡视器视角影像定位结果

表 2 中进一步将本文方法获取的地面影像空间定位结果与 RTK 实测坐标进行了对比。可以发现，本文方法所得到的位置误差在 0.3 米左右，可较好地满足巡视器定位的需求。本处定位误差的主要来源为石块空间坐标计算部分，由于相机标定和同名点坐标自动化量测过程存在着一定的不完善之处，空间前方交会计算得到的石块空间坐标与真实值略有差异，导致随后的空间后方交会部分产生误差；同时，本实验中 RTK 设备为手持，导致站点实测位置也存在着一定的人为误差。

表 2 本文方法计算结果与 RTK 对比

|  | 本文方法 | RTK | 误差 | 总误差 |
|---|---|---|---|---|
| X/m | 901386.6842 | 901386.5104 | 0.1738 | 0.3114 |
| Y/m | 3469782.1251 | 3469781.8667 | 0.2584 |  |

## 4 总结与展望

火星无人机的成功首飞为人类探索火星增加了新的方式。在未来的火星探测乃至其他具有大气条件的地外天体探测中（如 NASA 规划中的土卫六探测任务[13]），无人机也将成为一种重要的探测手段。本文提出了一种基于空地障碍分布模式匹配的火星巡视器空间定位方法，通过建立巡视器影像与无人机基准地图中的石块分布模式间的对应关系，获得巡视器在无人机基准地图中的空间位置。在同济大学月球与深空探测精密测绘综合实验场使用地面模拟数据获取的实验结果验证了本文方法的可行性。

与已有的基于特征点匹配的方法相比，本文方法在存在障碍特征但纹理较为贫乏的火星表面具有更好的空间定位能力。受时间所限，本研究仍有一定的可改进之处，主要包括：（1）更全面的定位精度评定。目前我们仅进行了单点定位实验，后续可进行多点序列定位的精度分析，以及本文方法与视觉里程计等方法的结果对比。（2）针对石块同名点匹配及分布模式匹配的算法还可考虑进一步优化。（3）考虑更多火星地表特征。本文只使用了石块作为空地影像匹配中的地物特征点，后续可考虑结合撞击坑、基岩等更多的目标，以提升该方法在不同地表环境下的适应性。（4）实际任务数据验证。本文所提出方法目前仅在地面模拟实验场进行了验证，拟对该方法在实际火星任务数据上的效果进行评估，以验证其应用于实际火星任务的可行性。